\documentclass[runningheads]{llncs}
\usepackage[T1]{fontenc}
\usepackage{graphicx,verbatim}
\usepackage{xcolor}
\usepackage{cite}
\usepackage{multirow}
\usepackage{booktabs}
\usepackage{amssymb}
\usepackage{caption}
\usepackage{subfigure}
\usepackage{threeparttable}
\usepackage{makecell}
\usepackage{amsmath}
\usepackage{amsfonts}
\usepackage{bbding}
\begin{document}
\title{Prompt-DAS: Annotation-Efficient Prompt Learning for Domain Adaptive Semantic Segmentation of Electron Microscopy Images}
\titlerunning{Prompt-DAS}
%
%
\author{Jiabao Chen\inst{1} \and
Shan Xiong\inst{1} \and
Jialin Peng\inst{1(}\Envelope\inst{)}}
\authorrunning{J. Chen et al.}
%
\institute{College of Computer Science and Technology, Huaqiao University, China 
\email{2004pjl@163.com}}
%
\maketitle              
\begin{abstract}
Domain adaptive segmentation (DAS) of numerous organelle instances from large-scale electron microscopy (EM)  is a promising way to enable annotation-efficient learning.  Inspired by SAM, we propose a promptable multitask framework, namely Prompt-DAS,  which is flexible enough to utilize any number of point prompts during the adaptation training stage and testing stage. Thus, with varying prompt configurations,  Prompt-DAS can perform unsupervised domain adaptation (UDA) and weakly supervised domain adaptation (WDA), as well as interactive segmentation during testing. Unlike the foundation model SAM, which necessitates a prompt for each individual object instance, Prompt-DAS is only trained on a small dataset and can utilize full points on all instances,  sparse points on partial instances, or even no points at all, facilitated by the incorporation of an auxiliary center-point detection task. Moreover,  a novel prompt-guided contrastive learning is proposed to enhance discriminative feature learning. Comprehensive experiments conducted on challenging benchmarks demonstrate the effectiveness of the proposed approach over existing UDA, WDA, and SAM-based approaches.
\end{abstract}
\section{Introduction}
Accurate semantic segmentation of subcellular organelles, e.g., mitochondria,  from various types of large-scale electron microscopy (EM) sequences is essential for cancer research and biology study\cite{neikirk2023call}. Although deep neural networks including convolutional neural networks \cite{ronneberger2015u} and
vision transformers (ViTs) \cite{dosovitskiy2020vit} have revolutionized the field of semantic segmentation for nearly all applications, including EM image segmentation \cite{peng2020unsupervised,wu2021uncertainty,huang2022domain,yin2023class,qiu2022wda}, the existing deep neural network models necessitate extensive pixel-wise annotations, involving expensive annotation budgets by experts. Furthermore, they typically show significant performance deterioration when applied directly to image datasets exhibiting different distributions.  This challenge is particularly relevant in the context of EM images, which experience significant domain shifts attributable to variations in microscopy techniques and tissue types. Manually annotating numerous organelle instances from large-scale EM images is time-consuming and labor-intensive. 

To reduce the burden of annotating each domain, we explore domain adaptation, which aims to reuse a well-trained model on a given source domain and adapt it to a target domain with a different distribution.  Although the unsupervised domain adaptation (UDA) completely assumes no annotation on the target domain, UDA methods still show relatively low performance on complicated tasks, which prohibits their practical usage.  To alleviate this issue, we leverage sparse points as \cite{qiu2024weakly} on the target domain as cheap weak labels to boost the segmentation performance with minimal annotation effort. In other words, we consider weakly supervised domain adaptation (WDA) with the same setting as WDA-Net \cite{qiu2024weakly}.  Compared to full point annotation for all object instances and pixel-wise annotation, the partial points demand substantially less time and expert knowledge \cite{qiu2024weakly}. Thus, annotating sparse points on a small number of object instances in EM images can be easily completed by non-experts. 

Recently, prompt-driven foundation models have shown remarkably strong generalization ability without training on specific targets, intriguing a trend toward more flexible segmentation paradigms. Notably, SAM \cite{kirillov2023segment}, which is pre-trained on billion-scale datasets of natural images, has demonstrated impressive performance on various segmentation tasks with points, boxes, or masks as user-generated prompts. These promptable segmentation models also pave the way for longstanding interactive segmentation, which can be responsive to user intention or progressively refine the segmentation, guided by the user input.   

However,  SAM  still has several limitations. First, SAM still struggles with domain shifts and usually shows low performance on medical image tasks, especially with point prompts, due to the lack of medical knowledge, ambiguous boundaries,  and complex shapes. To enhance the performance, several studies \cite{zhang2024improving,cheng2023sam,wu2023medical}, have proposed modifying or fine-tuning  SAM using medical data, such as SAM-Med2D \cite{cheng2023sam}, Med-SA \cite{wu2023medical}. Second, SAM lacks the functionality to segment all object instances of the same class without prompts on all instances,  making it particularly challenging to segment numerous organelle instances from EM images.
Third, SAM exhibits lower performance when using points as prompts, especially for medical images, as several studies \cite{xie2025self} have also shown.

Inspired by  SAM, we introduce Prompt-DAS, a promptable transformer model for domain-adaptive segmentation of EM images. Our model is flexible enough to utilize prompts during both the training and testing stages, offering advantages over previous UDA and WDA methods. To achieve a minimal annotation burden, we use sparse points as cost-effective prompts for the semantic segmentation of all object instances within the EM images.  Unlike SAM-based models, which require training on billion-scale datasets, we adapt a model trained on a source domain from scratch to a new target domain that already has sparse points available for the target training data and does not assume the availability of prompts during the testing stage.  However, when point prompts are available during the testing stage, the output of our method can further align with user intent. Moreover, our model conducts one-pass segmentation of many object instances with any point prompts, presenting an advantage over SAM. To enhance discriminative feature learning, we introduce a novel prompt-guided contrastive learning. 
Comprehensive experiments conducted on challenging benchmarks demonstrate the effectiveness of the
proposed approach.

\section{Method}
\textbf{Problem}. Suppose a source domain $\mathcal{D}^s=\{(x^s,y^s)\}$ with full pixel-wise labels $y^s$, and a target domain $\mathcal{D}^t=\{(x^t,\bar{c}^t)\}$ with point labels on centers of a few object instances. The binary dot label map $\bar{c}^t$  takes  1 only at the annotated sparse points. Additionally, the full dot label of a pixel-wise label $y$, denoted as $c$, has a corresponding density map  $d$ that is obtained through convolution with a Gaussian kernel $k_{\sigma}$, expressed as $d=k_\sigma*c$. Our objective is to develop a model that is flexible enough to perform UDA, WDA, as well as interactive versions of UDA and WDA. The model should also be flexible enough to be capable of effectively utilizing both training and testing prompts when provided.

\begin{figure}[t]
\centering
\includegraphics[width=1\textwidth]{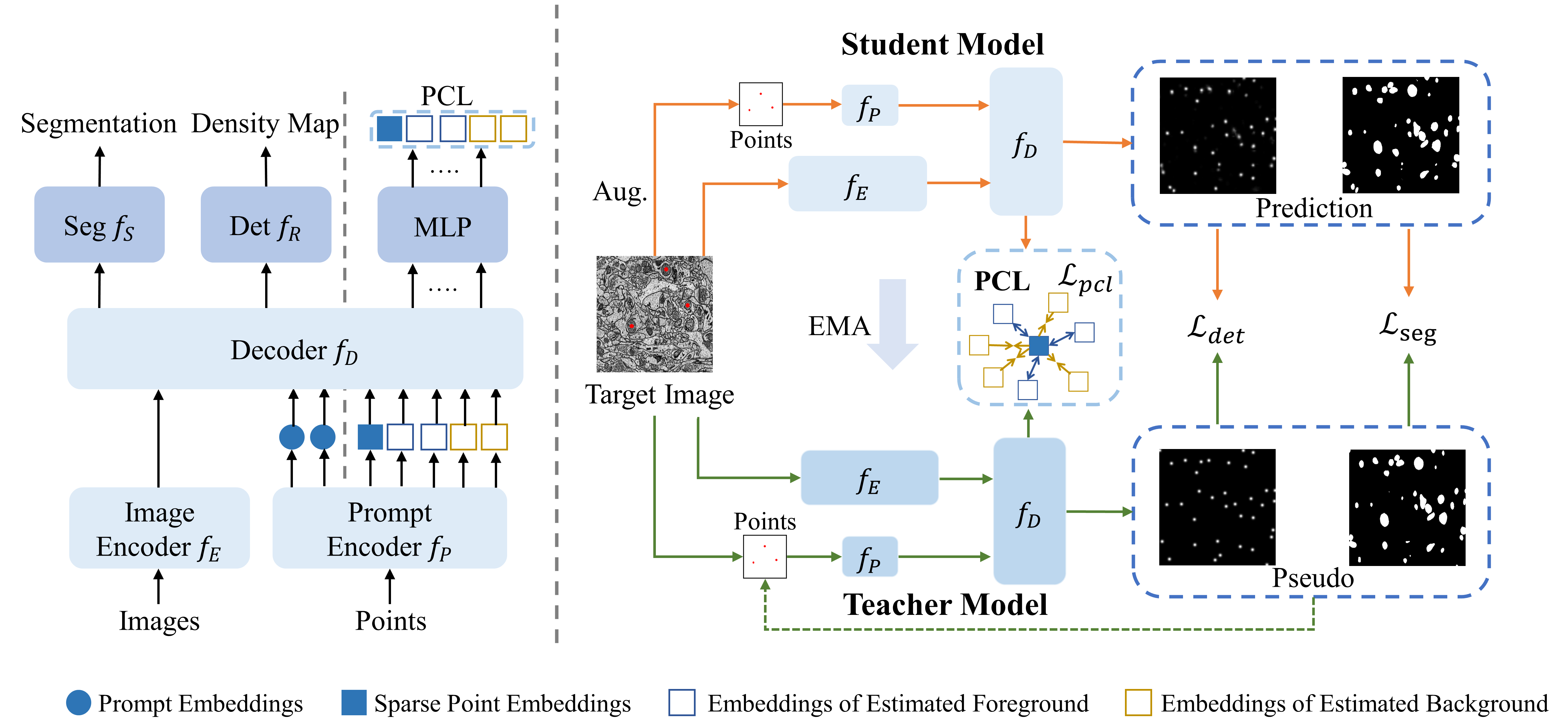}
\caption{Overview of our Prompt-DAS model for domain adaptive segmentation.} \label{fig1}
\end{figure}

\textbf{Overview}. Figure \ref{fig1} illustrates our Prompt-DAS, which encompasses an image encoder $f_E$, a point prompt encoder $f_P$  that processes $M\geq0$ points at once as inputs, a multitask decoder $f_D$ followed by a semantic segmentation head $f_S$, and a regression-based center-point detection head $f_R$.  In scenarios where $M=0$ points are provided on target training data, our model operates as  UDA, referred to as Prompt-DAS (0\%).  In scenarios where $M>0$ points are given on the target training data, our model conducts WDA learning. By default, we use 15\% sparse points, and our model is designated as  Prompt-DAS (15\%). When prompt points are provided during the testing phase, our model is capable of executing interactive UDA/WDA segmentation, denoted as Prompt-DAS+.

To tackle the issue of label scarcity on the target domain during domain adaptation learning, we conduct pseudo-label learning for both the segmentation and detection tasks under the mean-teacher framework \cite{NIPS2017_MT}. The output of the detection head $f_R$ is used to provide prompts for the segmentation task. Furthermore, the segmentation head $f_S$ is guided by a prompt-based contrastive loss, enhancing the discriminability of prompt embeddings.

\textbf{Promptable Detection.}
The proposed Prompt-DAS utilizes an auxiliary detection task to enhance the segmentation learning. 
The center point detection task is relatively easier than the dense segmentation task, particularly given sparse points as training prompts and partial supervision. While the joint learning of multiple tasks can implicitly boost the segmentation performance,   confident detection outputs are further employed to augment the ground-truth point prompts for the segmentation task. Following the teacher-student framework\cite{NIPS2017_MT}, pseudo-labels for the unlabeled regions are generated by selecting most highest local maxima points with a threshold from the predicted density map by the teacher model, which is updated by the exponential moving
average of the student network. Note that local maxima points can be identified through Non-Maxima Suppression. For the target data,  student network training is supervised by both the ground-truth sparse points and pseudo labels, and the $M$ sparse points are also used as training prompts. In the scenario of UDA, where there are no point annotations on the target domain, we use the estimated confident points from the prediction of the source model as the pseudo sparse points.
For the source data, ground truth center points are used as the supervision, and randomly sampled $n_s$ center points are used as training prompts. 
\begin{equation}
\mathcal{L}_{det}=\frac{1}{|\mathcal{D}^s|}\sum_{x^s}MSE(F_R(x^s),d^s)+\frac{1}{|\mathcal{D}^t|}\sum_{x^t}MSE(F_R(x^t),\hat{d}^t) \label{1}
\end{equation}
where $F_R$=$f_R\circ f_D\circ f_E$, $MSE$ represents mean square error loss, and $\hat{d}^t$ represents the density map generated by the target pseudo labels and ground truth points. 

\textbf{Promptable Segmentation.}
To alleviate label scarcity, we leverage pseudo-labeling in the teacher-student framework.  Thus, both ground truth source labels and target pseudo-labels are used to supervise the model training. 
\begin{equation}
\mathcal{L}_{seg} = \frac{1}{|D^s|} \sum_{x^s} CE(F_S(x^s), y^s) + \frac{1}{|D^t|} \sum_{x^t} CE(F_S(x^t), \hat{y}^t) \label{2}
\end{equation}
where $F_S$=$f_S\circ f_D\circ f_E$, $CE$ represents the standard cross-entropy loss, $\hat{y}^t$ represents the pseudo labels generated by the teacher model on the target domain. 

Similar to the detection head, we also use points as the segmentation training prompts. For the source domain, we use the $n_s$ points sampled for the detection task as the training prompts. Note that $n_s$ is a random number during training.  Since the target data only has a few points as the annotation, we propose to use both the estimated points from the detection output and ground-truth sparse points as training prompts to assist the segmentation training. The target point prompts are generated by selecting most highest local maxima points with a threshold from the predicted density map by the detection head.

\textbf{Prompt-guided Contrastive Learning (PCL).} To learn more discriminative embeddings during pseudo-label learning, we further introduce contrastive learning with the guidance of prompts, which can provide representative features to distinguish mitochondria instances from the background organelle. As shown in the left figure of Fig. \ref{fig1}, our contrastive learning aims to pull the feature embeddings of the estimated foreground points closer to that of the ground-truth sparse points while simultaneously pushing away from the foreground embeddings from the background embeddings. An MLP layer $\phi$ is utilized before conducting contrastive learning. Let $z^t = \phi(f_D(f_P(p^t)))$ denote the embedding derived from the target domain point $p^t$. We employ an attention mask mechanism following DN-DETR~\cite{li2022dn} to prevent information leakage from PCL.

Queries are generated from pixels identified as foreground exhibiting a sufficiently high confidence. Utilizing the pseudo-labels produced by the teacher model, we select three points from each instance with a confidence greater than $\delta_{f}$, resulting in $N^q$ foreground prompt embeddings $\{z^t_i\}_{i=1}^{N^q}$. Concurrently, we identify $N^n$ points with a confidence level below $\delta_{b}$, resulting in $N^n$ background prompt embeddings $\{\mu_k^b\}_{k=1}^{N^n}$. Since mitochondrial instances display high similarity, we employ the average embedding of sparse point prompts as the sparse prompt embedding $\mu^f$. The prompt-guided contrastive loss is defined as follows:
\begin{equation}
\mathcal{L}_{pcl} = -\sum_{i=1}^{N^q} \log \left[ \frac{\exp\left(\mu^f \cdot z^t_{i} /\tau) \right)}{\exp\left(\mu^f \cdot z^t_{i}/ \tau)  \right) +\sum_{k = 1}^{N^n} \exp\left( \mu_{k}^b \cdot z^t_{i} / \tau \right)} \right]\label{4}
\end{equation}
 In our experiments, we set $N^n=256$, $\delta_{f}=0.9$, and $\delta_{b}=0.1$.
\section{Experiments}
\textbf{Benchmark and Metrics.}
We evaluate the proposed method using the MitoEM dataset \cite{wei2020mitoem}, which is 3600 times larger than previous datasets and presents a greater challenge due to the wide diversity of mitochondria in terms of shape and density.
This dataset comprises two volumes of 30$\times$30$\times$30 $\mu m^3$  derived from the temporal lobe of an
adult human and the primary visual cortex of
an adult rat. The two datasets are named MitoEM-Human and MitoEM-Rat. 
The MitoEM-Human dataset contains significantly more mitochondria instances and a higher number of small mitochondria instances compared to the MitoEM-Rat dataset. Both datasets consist of 500 images of size 4096$\times$4096, where  400 images are allocated for training and 100 images for testing on each dataset. We consider the cross-domain segmentation between MitoEM-Human and MitoEM-Rat. We evaluate our method using the semantic-level Dice similarity coefficient (Dice) and the instance-level Aggregated Jaccard Index (AJI)~\cite{kumar2017dataset}, as well as Panoptic Quality (PQ)~\cite{Kirillov_2019_CVPR}.

\textbf{Implementation Details.}
We use the pre-trained ViT-S/8 with DINO~\cite{Caron_2021_ICCV} as our image encoder. Our decoder $f_D$ is similar to that of SAM but with mask attention and cross-attention to prevent information leakage. In contrast to SAM, our prompt encoder is a standard positional embedding. Our MLP is the same as the MLP of SAM and other standard transformers. More details can be found in our released code https://github.com/JiabaoChen1/Prompt-DAS.
The model is trained for 16k iterations with a batch size of 2, using the AdamW optimizer with an initial learning rate of $1 \times 10^{-5}$. The input images are subjected to random cropping, resulting in a size of 384$\times$384 pixels. We use a polynomial decay of power 0.9 to control the learning rate decay.  
\begin{table}[t]
\caption{Comparison results. We compare our Prompt-DAS under different settings with UDA, WDA, and SAM-based approaches.  For  WDA, sparse points on target training data are used as training prompts to achieve minimal annotation efforts. For interactive segmentation, indicated by a "+",  all center points of the testing data are used as testing prompts to fulfill the requirements of SAM.  } 
\centering
 \label{tab:1}
\begin{threeparttable}
\setlength{\tabcolsep}{0.3mm}
\begin{tabular}{lccccccccccc}
\toprule
\multirow{2}{*}{Methods} & \multicolumn{2}{c}{Prompts} & \multirow{2}{*}{Type} & ~ &\multicolumn{3}{c}{Human $\rightarrow$ Rat} & ~ 
&\multicolumn{3}{c}{Rat $\rightarrow$ Human}\\
\cmidrule{2-3}\cmidrule{6-8}\cmidrule{10-12}
& Training & Testing & ~ &~ &Dice  & AJI   & PQ  &~    
& Dice  & AJI   &PQ  \\
\midrule
SAM \cite{kirillov2023segment} & 
&  & \multirow{4}{*}{NoAdapt} & ~ & 32.0 & 14.3 & 30.0  &  
&  20.8 & 11.4 & 18.7 \\
SAM-Med2D \cite{cheng2023sam} & &  & ~ & & 15.9 & - & - & ~ 
& 22.5 &  - & - \\
Med-SA \cite{wu2023medical}\tnote{\dag}& &  & ~ & ~ & 75.5 & 56.6 & 27.5 & ~ 
& 72.4 & 55.0 & 33.4 \\
\textbf{Our Source Model} & ~ & ~ &  & ~
& 88.6& 76.7 & 68.7 & ~ 
&78.1& 62.8 & 55.5\\
\midrule
SAM\texttt{+} \cite{kirillov2023segment} &    & \checkmark &   & & 40.6 & 1.2 & 26.2& ~ 
& 40.3 & 4.6 &  26.6\\
SAM-Med2D\texttt{+} \cite{cheng2023sam} & & \checkmark & {NoAdapt} & & 72.6 & 55.6 &  39.7 & ~ 
&  78.1 &  61.2 & 42.2 \\
Med-SA\texttt{+} \cite{wu2023medical} & & \checkmark & (Interact.)  & & 86.2 & 70.2 & 59.9 & ~ 
& 83.8 & 68.1 & 59.0 \\
\textbf{Our Source Model\texttt{+}}& ~ & \checkmark & ~ & ~
& 89.8 & 78.8 & 74.1 & ~ 
& 87.6 & 76.0 & 70.6 \\
\midrule
DAMT-Net \cite{peng2020unsupervised} & ~ & ~ & \multirow{6}{*}{UDA} & ~& 88.7& 76.3&61.8 & ~
& 85.4& 72.3& 63.7 \\
UALR \cite{wu2021uncertainty} & ~ & ~ & ~ & ~& 86.3& 71.6&53.7 & ~
& 83.8& 69.7&60.0  \\
DA-ISC \cite{huang2022domain} (2.5D) & ~ & ~ & ~ & ~ & 88.6& 75.7&65.8 & ~
& 85.6& 72.7&63.8 \\
CAFA \cite{yin2023class} (2.5D) & ~ & ~ & ~ & ~ & 89.2& -&- & ~ 
& 86.6& -&- \\
WDA-Net (0\%)\cite{qiu2024weakly} & & & & & 88.2& 74.5&59.0 & ~
&85.5 &72.3 &60.6   \\
\textbf{Prompt-DAS (0\%)} &  & &  & &92.4 &82.2 &74.3  & 
& 88.0& 76.6&68.1\\
\midrule
WeSAM (15\%) \cite{zhang2024improving}\tnote{$\ddag$} & \checkmark &  & \multirow{3}{*}{WDA}& & 7.5 & 0.1 &  2.2 & ~ 
 & 3.6 & 1.3 & 1.6 \\
WDA-Net (15\%) \cite{qiu2024weakly} & \checkmark & &  & & 91.7& 80.7&74.0 & ~ 
& 88.7& 77.6&67.8 \\
\textbf{Prompt-DAS (15\%)} & \checkmark & ~ &  & &  93.3  &  83.6 & 74.5& ~ 
& 89.2 & 78.6 & 69.1 \\
\midrule
WeSAM (15\%)\texttt{+} \cite{zhang2024improving}\tnote{$\ddag$} & \checkmark & \checkmark &WDA  & & 89.9 & 79.6 & 73.9 & ~ 
& 82.3 & 66.0 & 65.5 \\
\textbf{Prompt-DAS(15\%)}\texttt{+} & \checkmark & \checkmark & (Interact.) & &  93.5  & 84.4 & 74.2& ~ 
& 90.8 & 81.5 & 72.3 \\
\midrule
Supervised model & & & Oracle & & 94.6& 86.4&  79.2& ~ & 92.6& 84.6& 75.8\\
\bottomrule
\end{tabular}
\begin{tablenotes}[flushleft] 
       \item \tnote{\dag} Fine-tuning using the source data
        \item \tnote{$\ddag$} Fine-tuning using the source data and  target data with 15\% sparse point labels
     \end{tablenotes} 
\end{threeparttable} 
\end{table}
For a fair comparison, the same data augmentations as those in WDA-Net are used. During the inference phase, we apply a sliding window with the same resolution as used during training. The implementation is conducted using PyTorch, and our model is trained for 6 hours on one RTX 4090 GPU with 24 GB of memory.

\textbf{Quantitative Evaluations.}  In Table \ref{tab:1}, we compare our Prompt-DAS model with six different types of models. 1) Methods without domain adaptation (NoAdapt):  SAM-based models without testing prompts, including SAM~\cite{kirillov2023segment}, SAM-Med2D~\cite{cheng2023sam}, and Med-SA$^\dag$~\cite{wu2023medical}, where $\dag$ means training using our source EM data; 2) NoAdapt with testing prompts for interactive segmentation: SAM\texttt{+}, SAM-Med2D\texttt{+}, and Med-SA\texttt{+};
3) SOTA 2D and 2.5D UDA methods, including DAMT-Net~\cite{peng2020unsupervised}, UALR~\cite{wu2021uncertainty}, DA-ISC~\cite{huang2022domain}, CAFA~\cite{yin2023class}, and WDA-Net (0\%)~\cite{qiu2024weakly}, and WeSAM~\cite{zhang2024improving} with fine-tuning on the source and target data; 4)  SOTA WDA methods, including  WDA-Net (15\%)~\cite{qiu2024weakly}), and WeSAM(15\%)$^\ddag$~\cite{zhang2024improving}, which use 15\% sparse points on the target training set and trained/finetued under the same setting as our Prompt-DAS(15\%); 5) Interactive version the WDA methods: WeSAM(15\%)\texttt{+}$^\ddag$, and Prompt-DAS(15\%)\texttt{+}; 6) The upperbound, which is the model fully supervised trained on the target domain. Additionally,  five settings of our model are in comparison, including our source model, our source model for interactive segmentation, our Prompt-DAS (0\%) for UDA segmentation, our Prompt-DAS (15\%) for WDA segmentation, and our Prompt-DAS (15\%)\texttt{+} for interactive WDA segmentation. It is noteworthy that, SAM, SAM-Med2D, Med-SA, and WeSAM are foundation models trained on billion-scale datasets,  with or without adaptation using large-scale medical data. In contrast, our Prompt-DAS model is trained from scratch. In the context of interactive segmentation,  all center points are used as testing prompts, as required by SAM; however, this setting is impractical for clinical usage. Conversely, during the testing stage, our model can effectively utilize sparse point prompts on partial object instances, enhancing its usability in real-world scenarios.

\begin{figure}[t]
    \centering
    \includegraphics[width=0.9\textwidth]{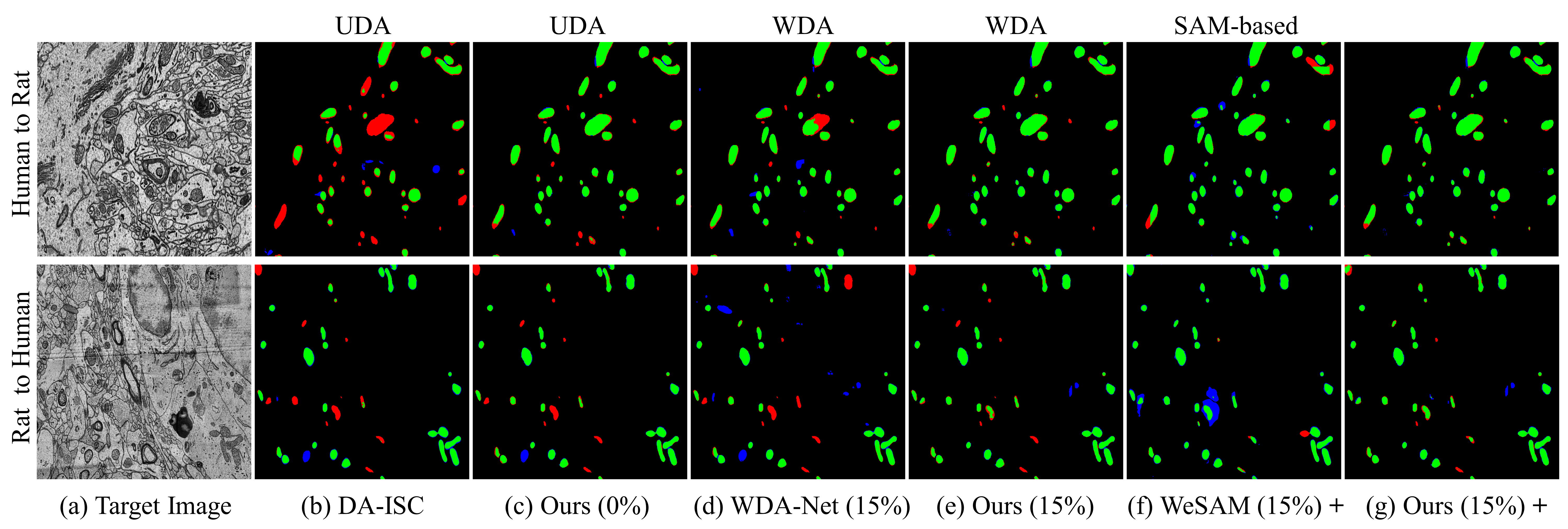}
    \caption{Qualitative comparison results on two adaptation tasks. Green: true positives; Red: false negatives; Blue: false positives.} 
    \label{fig2}
\end{figure}

Table \ref{tab:1} presents quantitative results on two domain adaptation tasks.  It is worth noting that the SAM and its medical versions show severely degraded performance both with and without testing prompts when directly applied to EM images. With fine-tuning using the EM data, the WeSAM demonstrates greatly improved performance over other SAM-based models. 
Our Prompt-DAS with 15\% sparse points as training prompts achieves the highest performance on both cross-domain segmentation tasks compared to all UDA, WDA, and SAM-based methods. Notably, our Prompt-DAS (0\%), the UDA version of our model, nearly outperforms all UDA, WDA, and SAM-based methods in comparison except the WDA-Net (15\%) on MitoEM-Rat $\rightarrow$  MitoEM-Human. With the inclusion of testing prompts, our   Prompt-DAS (15\%)+ achieves further performance gains, specifically a 1.6\% increase in  Dice over the Prompt-DAS (15\%) for MitoEM-Rat $\rightarrow$  MitoEM-Human. Compared to WDA-Net (15\%), our method demonstrates greater flexibility in alignment with human intention. Moreover,  our Prompt-DAS model exhibits performance that closely approaches the supervised upper bound, with only a minimal performance gap. Visual comparison results in Fig. \ref{fig2} further confirm the advantage of our method.

\textbf{Ablation study.}  In Table \ref{tab:2}, we evaluate the contributions of the key components of our approach: 1) Detection Pseudo-labeling; 2) Segmentation Pseudo-labeling; 3) Using sparse points as Training Prompts; 4) PCL: prompt-based contrastive learning.  As shown in  Table \ref{tab:2}, the base Model I, our source model, can be improved by adding pseudo-labeling-based detection or segmentation. By conducting multitask learning,  Model IV obtains a performance gain of 1.8\% in Dice over Model I. With training prompts, we further gain an improvement of 2.3\% in Dice over Model IV and obtain Model VI. With additional PCL, our full model further gains an improvement of 0.6\% in Dice and 0.4\% in PQ.

\begin{table}[t]
\caption{Ablation study on Human $\rightarrow$ Rat.}
\centering
\label{tab:2}
\setlength{\tabcolsep}{4pt}
\begin{tabular}{lccccccc}
\toprule
   & \multicolumn{2}{c}{Pseudo-labeling}  & Training  & \multirow{2}{*}{PCL} &\multirow{2}{*}{Dice (\%)} &\multirow{2}{*}{PQ (\%)}\\
   \cline{2-3} &  Detection& Segmentation & Prompts &  & &\\
\midrule
\uppercase\expandafter{\romannumeral1}  &~  & ~         & ~          & ~ &  88.6 & 68.7 \\
\uppercase\expandafter{\romannumeral2}  &\checkmark & ~          & ~          & ~ &  89.2 & 70.4\\
\uppercase\expandafter{\romannumeral3}   & ~          &\checkmark          & ~ & ~ & 89.5 & 70.0 \\
\uppercase\expandafter{\romannumeral4}     &\checkmark &\checkmark          & ~& ~ & 90.4 & 71.8 \\
\uppercase\expandafter{\romannumeral5}     &\checkmark &~          & \checkmark& ~ & 89.8 & 71.7 \\
\uppercase\expandafter{\romannumeral6}     &\checkmark &\checkmark   & \checkmark & ~ & 92.7 & 74.1 \\
Full     &\checkmark &\checkmark   & \checkmark & \checkmark & 93.3 & 74.5 \\
\bottomrule
\end{tabular}
\end{table}

\begin{table}[t]
\centering
 \setlength{\tabcolsep}{1.5mm}
 \renewcommand\arraystretch{1.1}
\caption{The impact of testing prompt amount on interactive segmentation.}
\label{tab:3}
\begin{tabular}{cccccc}
\toprule
\multirow{2}{*}{\thead{Testing Prompts \\ (Points)}}&\multicolumn{2}{c}{Human $\rightarrow$  Rat} & ~ & \multicolumn{2}{c}{Rat $\rightarrow$ Human}\\
\cline{2-3}\cline{5-6}
 & Dice (\%)  & PQ (\%)  & ~ & Dice (\%)  & PQ (\%)\\
 \midrule
0 & 93.3& 74.5 & ~ & 89.2 & 69.1\\
15\%  & 93.5& 74.2 & ~ & 90.0 & 69.8\\
50\%  & 93.5& 74.2 & ~ & 90.4 & 70.9\\
100\% & 93.5& 74.2 & ~ & 90.8 & 72.3\\
\bottomrule
\end{tabular}
\end{table}

\textbf{Influence of testing prompts.} Compared to SAM, our model is flexible enough to utilize partial center points as testing prompts. Table \ref{tab:3} presents the influence of testing prompts for interactive segmentation. For Rat $\rightarrow$  Human, our model can gain improved performance with more point prompts. However, for Human $\rightarrow$  Rat,  adding point prompts does not improve the performance. The main reason is that our model's performance is already very close to the supervised upper bound, and there is a minimal number of false negatives. Moreover, our model can achieve similar performance with only 15\% partial points, taking less than 1/5 of the annotation time of full points.
\section{Conclusion}
In this study, we develop a promptable transformer model for domain adaptive segmentation of EM images, which can conduct UDA, WDA, and interactive segmentation with various prompting configurations, including sparse points. Our model augments the segmentation task with a detection task, which can significantly alleviate label scarcity and generate pseudo-prompts for the segmentation. Furthermore, the segmentation is guided by a prompt-based contrastive loss, enhancing the discriminability of prompt embeddings.  Comprehensive experiments conducted on challenging benchmarks demonstrate the SOTA performance of our approach. The limitation of our model is its requirement for source data and labels for training. In future studies, we will consider a source-free setting.
\subsubsection{\ackname} This work was partially supported by the NSFC   (No. 1247011276) and  Xiamen Natural Science Foundation
(No. 3502Z202373042).
\subsubsection{\discintname}
The authors have no competing interests to declare that are relevant to the content of this article.

%
%
%
%
\bibliographystyle{splncs04}
\bibliography{Paper-1547}

\begin{thebibliography}{10}
\providecommand{\url}[1]{\texttt{#1}}
\providecommand{\urlprefix}{URL }
\providecommand{\doi}[1]{https://doi.org/#1}

\bibitem{Caron_2021_ICCV}
Caron, M., Touvron, H., Misra, I., J\'egou, H., Mairal, J., Bojanowski, P.,
  Joulin, A.: Emerging properties in self-supervised vision transformers. In:
  Proceedings of the IEEE/CVF International Conference on Computer Vision. pp.
  9650--9660 (2021)

\bibitem{cheng2023sam}
Cheng, J., Ye, J., Deng, Z., Chen, J., Li, T., Wang, H., Su, Y., Huang, Z.,
  Chen, J., Jiang, L., et~al.: Sam-med2d. arXiv preprint arXiv:2308.16184
  (2023)

\bibitem{dosovitskiy2020vit}
Dosovitskiy, A., Beyer, L., Kolesnikov, A., Weissenborn, D., Zhai, X.,
  Unterthiner, T., Dehghani, M., Minderer, M., Heigold, G., Gelly, S.,
  Uszkoreit, J., Houlsby, N.: An image is worth 16x16 words: Transformers for
  image recognition at scale. In: International Conference on Learning
  Representations (2021)

\bibitem{huang2022domain}
Huang, W., Liu, X., Cheng, Z., Zhang, Y., Xiong, Z.: Domain adaptive
  mitochondria segmentation via enforcing inter-section consistency. In:
  International Conference on Medical Image Computing and Computer-Assisted
  Intervention. pp. 89--98 (2022)

\bibitem{Kirillov_2019_CVPR}
Kirillov, A., He, K., Girshick, R., Rother, C., Dollar, P.: Panoptic
  segmentation. In: Proceedings of the IEEE/CVF Conference on Computer Vision
  and Pattern Recognition. pp. 9404--9413 (2019)

\bibitem{kirillov2023segment}
Kirillov, A., Mintun, E., Ravi, N., Mao, H., Rolland, C., Gustafson, L., Xiao,
  T., Whitehead, S., Berg, A.C., Lo, W.Y., et~al.: Segment anything. In:
  Proceedings of the IEEE/CVF International Conference on Computer Vision. pp.
  4015--4026 (2023)

\bibitem{kumar2017dataset}
Kumar, N., Verma, R., Sharma, S., Bhargava, S., Vahadane, A., Sethi, A.: A
  dataset and a technique for generalized nuclear segmentation for
  computational pathology. IEEE Transactions on Medical Imaging
  \textbf{36}(7),  1550--1560 (2017)

\bibitem{li2022dn}
Li, F., Zhang, H., Liu, S., Guo, J., Ni, L.M., Zhang, L.: Dn-detr: Accelerate
  detr training by introducing query denoising. In: Proceedings of the IEEE/CVF
  Conference on Computer Vision and Pattern Recognition. pp. 13619--13627
  (2022)

\bibitem{neikirk2023call}
Neikirk, K., Lopez, E.G., Marshall, A.G., Alghanem, A., Krystofiak, E., Kula,
  B., Smith, N., Shao, J., Katti, P., Hinton~Jr, A.: Call to action to properly
  utilize electron microscopy to measure organelles to monitor disease.
  European Journal of Cell Biology  \textbf{102}(4),  151365 (2023)

\bibitem{peng2020unsupervised}
Peng, J., Yi, J., Yuan, Z.: Unsupervised mitochondria segmentation in em images
  via domain adaptive multi-task learning. IEEE Journal of Selected Topics in
  Signal Processing  \textbf{14}(6),  1199--1209 (2020)

\bibitem{qiu2024weakly}
Qiu, D., Xiong, S., Yi, J., Peng, J.: Weakly-supervised cross-domain
  segmentation of electron microscopy with sparse point annotation. IEEE
  Transactions on Big Data  \textbf{11}(2),  359--371 (2025)

\bibitem{qiu2022wda}
Qiu, D., Yi, J., Peng, J.: Wda-net: Weakly-supervised domain adaptive
  segmentation of electron microscopy. In: IEEE International Conference on
  Bioinformatics and Biomedicine. pp. 1132--1137 (2022)

\bibitem{ronneberger2015u}
Ronneberger, O., Fischer, P., Brox, T.: U-net: Convolutional networks for
  biomedical image segmentation. In: International Conference on Medical Image
  Computing and Computer-Assisted Intervention. pp. 234--241 (2015)

\bibitem{NIPS2017_MT}
Tarvainen, A., Valpola, H.: Mean teachers are better role models:
  Weight-averaged consistency targets improve semi-supervised deep learning
  results. In: Advances in Neural Information Processing Systems. vol.~30
  (2017)

\bibitem{wei2020mitoem}
Wei, D., Lin, Z., Franco-Barranco, D., Wendt, N., Liu, X., Yin, W., Huang, X.,
  Gupta, A., Jang, W.D., Wang, X., et~al.: Mitoem dataset: Large-scale 3d
  mitochondria instance segmentation from em images. In: International
  Conference on Medical Image Computing and Computer-Assisted Intervention. pp.
  66--76 (2020)

\bibitem{wu2023medical}
Wu, J., Ji, W., Liu, Y., Fu, H., Xu, M., Xu, Y., Jin, Y.: Medical sam adapter:
  Adapting segment anything model for medical image segmentation. arXiv
  preprint arXiv:2304.12620  (2023)

\bibitem{wu2021uncertainty}
Wu, S., Chen, C., Xiong, Z., Chen, X., Sun, X.: Uncertainty-aware label
  rectification for domain adaptive mitochondria segmentation. In:
  International Conference on Medical Image Computing and Computer Assisted
  Intervention. pp. 191--200 (2021)

\bibitem{xie2025self}
Xie, B., Tang, H., Cai, D., Yan, Y., Agam, G.: Self-prompt sam: Medical image
  segmentation via automatic prompt sam adaptation. arXiv preprint
  arXiv:2502.00630  (2025)

\bibitem{yin2023class}
Yin, D., Huang, W., Xiong, Z., Chen, X.: Class-aware feature alignment for
  domain adaptative mitochondria segmentation. In: International Conference on
  Medical Image Computing and Computer-Assisted Intervention. pp. 238--248
  (2023)

\bibitem{zhang2024improving}
Zhang, H., Su, Y., Xu, X., Jia, K.: Improving the generalization of
  segmentation foundation model under distribution shift via weakly supervised
  adaptation. In: Proceedings of the IEEE/CVF Conference on Computer Vision and
  Pattern Recognition. pp. 23385--23395 (2024)

\end{thebibliography}
\end{document}